\documentclass[letterpaper,journal]{IEEEtran}
\usepackage{amsmath,amsfonts}
\usepackage{algorithmic}
\usepackage{algorithm}
\usepackage{array}
\usepackage[caption=false,font=normalsize,labelfont=sf,textfont=sf]{subfig}
\usepackage{textcomp}
\usepackage{stfloats}
\usepackage{url}
\usepackage{verbatim}
\usepackage{amsthm}

\newtheorem{theorem}{Theorem}[section]
\newcommand{\tens}[1]{%
  \mathbin{\mathop{\otimes}\limits_{#1}}%
}

\usepackage{graphicx}
\usepackage{cite}
\usepackage{booktabs}
\usepackage{bibentry}
\newtheorem{assumption}{Assumption}
\begin{document}

\title{IntOPE: Off-Policy Evaluation in the Presence of Interference}

\author{%
    Yuqi Bai,
    Ziyu Zhao,
    Chenxin Lyu,
    Minqin Zhu,
    Kun Kuang
\thanks{Yuqi Bai is with the John A. Paulson School of Engineering and Applied Sciences, Harvard University, Cambridge, MA, USA (email: yuqi\_bai@g.harvard.edu).}%
\thanks{Ziyu Zhao, Minqin Zhu, and Kun Kuang are with the College of Computer Science, Zhejiang University, Hangzhou, China (email: benzhao.styx@gmail.com; minqinzhu@zju.edu.cn; kunkuang@zju.edu.cn).}%
\thanks{Chenxin Lyu is with the University of Waterloo, Waterloo, ON, Canada (email: c33lyu@uwaterloo.ca).}%

    \thanks{Manuscript received April 19, 2021; revised August 16, 2021.}
}


\markboth{Journal of \LaTeX\ Class Files,~Vol.~14, No.~8, August~2021}%
{Shell \MakeLowercase{\textit{et al.}}: A Sample Article Using IEEEtran.cls for IEEE Journals}

\IEEEpubid{0000--0000/00\$00.00~\copyright~2021 IEEE}

\maketitle

\begin{abstract}
Off-Policy Evaluation (OPE) is employed to assess the potential impact of a hypothetical policy using logged contextual bandit feedback, which is crucial in areas such as personalized medicine and recommender systems, where online interactions are associated with significant risks and costs. Traditionally, OPE methods rely on the Stable Unit Treatment Value Assumption (SUTVA), which assumes that the reward for any given individual is unaffected by the actions of others. However, this assumption often fails in real-world scenarios due to the presence of interference, where an individual's reward is affected not just by their own actions but also by the actions of their peers. This realization reveals significant limitations of existing OPE methods in real-world applications. To address this limitation, we propose IntIPW, an IPW-style estimator that extends the Inverse Probability Weighting (IPW) framework by integrating marginalized importance weights to account for both individual actions and the influence of adjacent entities. Extensive experiments are conducted on both synthetic and real-world data to demonstrate the effectiveness of the proposed IntIPW method.
\end{abstract}


\section{Introduction}
\IEEEPARstart{T}{he} context bandit process is important across a range of real-world applications, including precision medicine~\cite{10.1145/3383313.3412244}, recommender systems~\cite{Li_2011}, and advertising~\cite{Li_2011}. In these interactive systems, a logging policy observes user contexts, executes actions (treatments), and collects rewards (outcomes), thereby generating a substantial volume of logs. These logs, resulting from online interactions, provide critical data for off-policy evaluation (OPE). OPE is designed to assess the performance of new policies without resorting to A/B testing~\cite{lewis2008does}, thereby reducing high costs and potential ethical concerns. 

\begin{figure}[ht]
\centering
\includegraphics[width=1\linewidth]{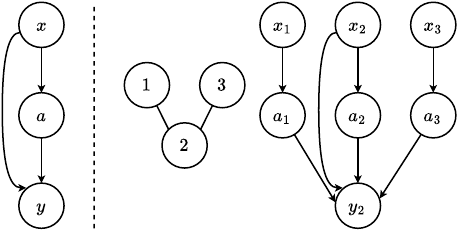}
\caption{Example of a small social network. $x$, $a$ and $y$ are contexts, actions, and rewards respectively. Left: causal graph of common SUTVA setting. Right: the network connection topology and causal graph of our setting. For simplicity, $y_1$ and $y_3$ are not shown.}
\label{fig:motivating_example}
\end{figure}

As interest in the OPE problem grows, numerous estimators have been developed to use offline log data for assessing new policies. Direct Method (DM)~\cite{beygelzimer2016offset} learns a predictive reward function from the log data. Inverse Probability Weighting (IPW)~\cite{Horvitz1952AGO, DBLP:journals/corr/abs-1003-0120} assigns sample weights based on the propensities of the behavior and evaluation policies. Doubly Robust (DR)~\cite{dudik2011doubly} combines DM and IPW.
These estimators have demonstrated effectiveness across various domains~\cite{saito2022offpolicy, 10.1145/3543507.3583448, kallus2018policy}. The effectiveness of these methods relies on the Stable Unit Treatment Value Assumption (SUTVA), which assumes that the reward for any given unit is influenced solely by its context and the action it takes. However, the SUTVA is often violated in many real-world scenarios involving a social network since peers are not independent~\cite{ma2021causal}.

In a social network, interference occurs when a unit’s reward is influenced not only by its own action but also by the actions of its social contacts, a phenomenon also known as peer effects~\cite{ogburn2022causal}. For instance, in epidemiology, deciding whether to vaccinate a person against an infectious disease can affect the health status of those around them~\cite{Fine2011HerdImmunity}; in recommender systems, recommending a product to a person can also indirectly influence the purchasing intentions of others within their social circle~\cite{Social_Circle} through opinion propagation; in advertising, an ad's exposure may directly affect a user's purchase behavior and indirectly affect others in the social network through the acquisition behavior of others~\cite{parshakov2020spillover}. However, current studies on OPE neglect the peer effects which may cause these estimators to fail completely. Therefore, how to evaluate a new policy from logging data in the presence of interference remains an open problem. 

\IEEEpubidadjcol

In this paper, we study the problem of off-policy evaluation in social networks in the presence of interference. As shown in the left half of Fig.\ref{fig:motivating_example}, prior studies on OPE mainly adhered to the SUTVA. However, in real-world scenarios, as shown on the right side of Fig.\ref{fig:motivating_example}, there often exists a social network in which people with social relationships influence each other, thus violating the SUTVA. Specifically, for each individual, the reward is influenced not only by the action they take but also by the actions of other individuals in their social relationships.

To bridge this gap, in this paper, we propose an IPW-style estimator termed IntIPW, which uses marginalized importance weights to account for both individual actions and the effects of adjacent entities. Specifically, IntIPW simplifies the process of weight estimation into a binary classification task. This is achieved by first assigning unique labels to context-action pairs from the logging policy and the evaluation policy. Following this, a classifier that employs a Graph Convolutional Network (GCN)~\cite{10.1145/3336191.3371816, Chu_2021, kipf2017semisupervised} is trained to differentiate the sample distributions from different policies. Ultimately, the logged samples are assigned weights based on the classifier's predictions, thereby presenting a new approach within the IPW framework to effectively bridge the existing gap. Empirical studies conducted on two synthetic datasets and two real-world datasets demonstrate the effectiveness of the proposed method in the presence of interference.

The main contributions of this paper are as follows:
\begin{itemize}
    \item We investigate a new problem of off-policy evaluation in social networks in the presence of interference. 
    \item We propose IntIPW estimator, which uses marginalized importance weights to account for both individual actions and adjacent entities' effects.
    \item Empirical results on both synthetic and real-world datasets show that our estimator outperforms other baselines in OPE in the presence of interference.
\end{itemize}

\section{Related Works}
\subsection{Off-Policy Evaluation} Common OPE estimators mainly include the Direct Method (DM)~\cite{beygelzimer2016offset, DBLP:journals/corr/abs-1003-0120} that directly learns the reward function, the Inverse Probability Weighting (IPW)~\cite{Horvitz1952AGO} method which weights samples based on the inverse of the propensity score, and the Doubly Robust (DR)~\cite{dudik2011doubly} method which combines both. Among these, the IPW method is the most common.

In practical scenarios, when the behavior and evaluation policies are very dissimilar, the estimates from IPW tend to be poor. To mitigate this issue,~\cite{NEURIPS2021_4476b929} introduce Sub-Gaussian IPW (SGIPW), which modifies the vanilla weights by applying power mean. This method provides robustness against outliers and extreme values in data distributions.

When the number of actions is large, IPW-based estimators can suffer from significant bias and variance because ``full support'', a key theoretical requirement of IPW, is violated. This happens when the probability of some actions is zero in some contexts~\cite{Sachdeva_2020}. To solve this problem,~\cite{Sachdeva_2020} identify three approaches: restricting the action space, reward extrapolation, and restricting the policy space. Alternatively,~\cite{saito2022offpolicy} propose  Marginalized IPW method, which utilizes action embeddings. In addition,~\cite{10.1145/3543507.3583448} propose GroupIPW estimator, whose main idea is action grouping. 

Another issue that needs to be addressed is that the vanilla IPW estimator is not applicable when the action space is continuous. To resolve this problem,~\cite{kallus2018policy} introduce a new IPW estimator that generalizes IPW and DR to continuous action spaces by using a kernel function.

Similar to the interference problem studied in this work, \cite{zhang2024individualizedpolicyevaluationlearning} studied the problem of OPE under clustered network interference. In this scenario,  In their scenario, individuals are grouped into clusters where the treatment received by one person impacts others within the same group. They introduce the additive IPW estimator alongside a policy learning method to address the OPE problem in this setting.



\subsection{Interference} In recent years, there has been a considerable amount of literature studying network interference from various perspectives. A large portion of such work has focused on studying average treatment effects in observational data of homogeneous networks, where the influence of each unit on every neighbor is assumed to be uniform~\cite{liu2016inverse, ogburn2022causal, Sofrygin2016SemiParametricEA, doi:10.1080/01621459.2020.1811098}. In terms of individual treatment effects,~\cite{ma2021causal} research causal effect estimation under general network interference using Graph Neural Networks.~\cite{ma2022learning} investigate high-order interference modeling and proposed a new causality learning framework powered by hypergraph neural networks.

In the real world, interference is often heterogeneous, meaning that the influence of a node on different neighbors varies~\cite{qu2022efficient}. In this context,~\cite{song2019session} propose a recommender system for online communities based on a dynamic-graph-attention neural network. Similarly, aiming at providing a general method for improving recommender systems by including social network information,~\cite{ma2011recommender} propose a matrix factorization framework with social regularization.

\section{PRELIMINARIES}
In this section, we begin by outlining a formal definition of the contextual bandit framework when interference is present. Subsequently, we enumerate several traditional methods for off-policy evaluation and explore their constraints where interference occurs.

\subsection{Problem Setup}
In this paper, we explore a scenario that extends beyond the conventional contextual bandit framework by including the influence of social networks on off-policy evaluation. The dataset includes four key components: $\mathbf{x} \in \mathcal{X} \subseteq \mathbb{R}^p$ are $p$-dimensional vectors that represent the contexts of units; $\mathbf{a} \in \mathcal{A} $ are the actions taken by units, where $\mathcal{A} = \{1, 2, \ldots, D\}\ $ is a discrete and finite action space of size \(D\); $\mathbf{r} \in (-\infty, \infty)$ is the rewards obtained by units; $A$ is the adjacency matrix representing the link information of the social network, where $A[i,j] = A[j,i] = 1$ if units $i$ and $j$ are neighbors, and $A[i,j] = A[j,i] = 0$ otherwise.

The process of dataset collection is:
\begin{itemize}
    \item The context of each unit $x_i$ is sampled from the population distribution $x_i  \sim p(x)$. 
    \item  The action taken by each unit $a_i$ is drawn by the behavior policy $\pi_0$, which is stochastic, conditional on the context of the unit. That is, $a_i \sim \pi_0(a |x_i) $. 
    \item The reward received by each unit $r_i$ depends on its context, action, and actions of its neighbors. That is, $r_i \sim p(r|x_i, a_i, a_{i\mathcal{N}})$, where $a_{\mathcal{N}}$ is the aggregation of neighbors' actions. 
\end{itemize}
To summarize, the dataset follows such a distribution:
\begin{equation}
\mathcal{D} := A, \ \{ (x_i, a_i, r_i) \}_{i=1}^{n} \sim p(x) \tens{} \pi_0(a | x)\tens{} p(r | x, a, a_N).
\label{eq:data_definition}
\end{equation}

To ensure that the historical dataset can be utilized for an unbiased estimation of the effectiveness of the evaluation policy, traditional estimators usually require the SUTVA. However, the SUTVA does not hold in our setting because the reward of a unit is influenced by the actions of 1-hop neighbors. Instead, following~\cite{10.1145/3511808.3557311, zhao2024learning}, we assume that the interference has the Markov property:
\begin{assumption}
\label{assumption:markov}
(Markov) The potential reward for any unit is influenced by its own context, its action, and the action of its neighbors only:
\begin{equation}
r_i \not\!\perp\!\!\!\perp a_{i-\mathcal{N}} \mid x_i, a_i, a_{i\mathcal{N}},
\label{eq:interference}
\end{equation}
where \(a_{i-\mathcal{N}}\) is the actions of the non-neighbor units of unit \(i\).

\end{assumption}

In addition, adequate overlap in the action distributions of the evaluation policy and the behavior policy is required~\cite{Sachdeva_2020}. Formally, networked common support is defined as: 
\begin{assumption}
\label{assumption:common_support}
(Networked Common Support) The behavior policy $\pi_0$ is said to have common support for the evaluation policy $\pi$ if 
\begin{equation}
p(a_i, a_{i\mathcal{N}} | x_i, A, \pi) > 0 \rightarrow p(a_i, a_{i\mathcal{N}} | x_i, A, \pi_0) > 0
\label{eq:common_support}
\end{equation}
for all $a \in \mathcal{A}$, $x \in \mathcal{X}$, and unit $i$. 
\end{assumption}

Our objective is to estimate the effectiveness of the evaluation policy $ \pi $ with only $ \mathcal{D} $ given:
\begin{equation}
V(\pi) := \mathbb{E}_{p(x)\pi(a|x)p(a_{N}|A,x,\pi)p(r|x,a,a_{N})}[r].
\label{eq:evaluation_policy}
\end{equation}
To achieve this, we need to develop an estimator $\hat{V}$. The accuracy of $\hat{V}$ can be measured by the MSE~\cite{saito2022offpolicy}:
\begin{align}
\text{MSE}(\hat{V}(\pi)) &:= \mathbb{E}_\mathcal{D}\left[\left(V(\pi) - \hat{V}(\pi; \mathcal{D})\right)^2\right] \notag \\
&= \text{Bias}^2\left(\hat{V}(\pi)\right) + \text{Var}_\mathcal{D}\left(\hat{V}(\pi; \mathcal{D})\right). \label{eq:mse_decomposed}
\end{align}
\subsection{Existing Estimators}
Next, we introduce some traditional OPE estimators and analyze their limitations in social networks with interference. \\
\textbf{Direct Method (DM).} DM estimator involves learning a reward function $ \hat{r}(x, a) $ from historical data. It then uses this reward function to directly estimate the reward of the evaluation policy. 
\begin{equation}
\hat{V}_{DM} = \frac{1}{n} \sum_{i=1}^{n} \sum_{a \in \mathcal{A}} \pi(a|x_i) \hat{r}(x_i, a).
\label{eq:DM}
\end{equation}
\textbf{Inverse Probability Weighting (IPW).} This method is also known as Inverse Propensity Score or Importance Sampling. In IPW estimator, the inverse of the probabilities of selecting an action is used as a weight to reweight the rewards of samples in the dataset.
\begin{equation}
\hat{V}_{IPW} = \frac{1}{n} \sum_{i=1}^{n} \frac{\pi(a_i|x_i)}{\pi_0(a_i|x_i)} r_i.
\label{eq:IPW}
\end{equation}
$ w(x_i, a_i) := \frac{\pi(a_i|x_i)}{\pi_0(a_i|x_i)} $ is called the (vanilla) importance weight~\cite{saito2022offpolicy}. \\
\textbf{Self-Normalized Inverse Probability Weighting (SNIPW).} Based on the vanilla IPW, this method self-normalizes the weights. By doing so, the problem that IPW suffers from propensity overfitting when the mean of weights deviates from its expected value of 1 is improved~\cite{NIPS2015_39027dfa}. This adjustment also reduces the variance. 
\begin{equation}
\hat{V}_{SNIPW} = \frac{\sum_{i=1}^{n} \frac{\pi(a_i|x_i)}{\pi_0(a_i|x_i)} r_i}{\sum_{i=1}^{n} \frac{\pi(a_i|x_i)}{\pi_0(a_i|x_i)}}.
\label{eq:SNIPW}
\end{equation}
\textbf{Doubly Robust (DR).} Doubly robust estimator is a combination of the DM estimator and the IPW estimator. The term ``doubly robust" refers to the estimator's key feature that it is unbiased when either the propensity scores or estimated rewards are correct.
\begin{equation}
\begin{aligned}
\hat{V}_{DR} = \frac{1}{n} \sum_{i=1}^{n} & \left[ \sum_{a \in \mathcal{A}} \pi(a|x_i) \hat{f}(x_i, a) \right. \\
& \left. + \frac{\pi(a_i|x_i)}{\pi_0(a_i|x_i)} \left(r_i - \hat{f}(x_i, a_i)\right) \right].
\label{eq:dr}
\end{aligned}
\end{equation}
\indent These estimators are not suitable for social networks where interference exists, because they all require the SUTVA, which assumes the potential reward for any unit is not influenced by the action assignment of other units, to ensure performance.

For DM, it may exhibit significant bias when the discrepancy in the distribution of actions between the behavior policy and the evaluation policy is large~\cite{dudik2011doubly}. Moreover, in the presence of interference, DM becomes even more biased because instead of learning a reward function \( \hat{r}(x, a) \), it should learn \( \hat{r}(x, a, a_{\mathcal{N}}) \).

Moreover, the vanilla IPW is also biased in the presence of interference because the vanilla importance weights do not take into account the effect of neighbors' actions. Ideally, the weights should consider both the units' own actions and the aggregation of neighbors' actions:
\begin{equation}
w(a_i, a_{i\mathcal{N}}, x_i) = \frac{p(a_i, a_{i\mathcal{N}} |x_i, A, \pi)}{p(a_i, a_{i\mathcal{N}}|x_i, A, \pi_0)}.
\label{eq:ideal_ipw_weight}
\end{equation}
Correspondingly, an ideal unbiased IPW estimator would be: 
\begin{equation}
\hat{V}_{\text{ideal-IPW}} = \frac{1}{n} \sum_{i=1}^{n} \frac{p(a_i, a_{i\mathcal{N}} | x_i, A, \pi)}{p(a_i, a_{i\mathcal{N}} | x_i, A, \pi_0)}  r_i.
\label{eq:ideal_ipw}
\end{equation}

\section{Algorithm}


In this section, we provide a detailed description of our IntIPW algorithm that aims to solve the OPE problem in social networks in the presence of interference. Our method uses marginalized importance weights to account for both individual actions and adjacent entities' effects. 

In the presence of interference, not only the vanilla IPW but also all estimators that fail to account for the influence of neighbors' actions are biased. Therefore, we need to propose a new estimator to address the OPE problem in the presence of interference.
The ideal IPW estimator Eq.\ref{eq:ideal_ipw} is quite intuitive, but without additional assumptions~\cite{li2022random}, it is difficult to determine how much a unit is affected by the actions of its neighbors. Alternatively, inspired by~\cite{kipf2017semisupervised}, we conceived the idea of representing the adjacency information with a graph, and then using a GCN to aggregate units' own actions with the actions of neighbors. Subsequently, we convert the problem of assigning weights into a binary classification problem~\cite{sondhi2020balanced}. Building on this foundation, we propose a supervised learning approach to address the challenge of OPE in the presence of interference. 

First, we assign a label $ L=0 $ to the historical samples (generated by the behavior policy) $ \{\mathbf{x}, \mathbf{a_0}, A\} $ and a label $ L=1 $ to the samples generated by the evaluation policy $ \{\mathbf{x}, \mathbf{a}, A\} $. These two datasets have the same state distribution $\mathbf{x} \sim p(x)$ and size. The only difference is their action distributions $\mathbf{a_0} \sim \pi_0(a | x)$ and $\mathbf{a} \sim \pi(a | x)$.

With the labeled datasets, we want to learn joint representations for each sample. To do so, we use an embedding function \(e\) to map actions into the semantic space, and concatenate the contexts and the action embeddings for the labeled datasets as \(\textbf{h}_0\) and \(\textbf{h}_1\):
\begin{align}
\textbf{h}_0 &= [\mathbf{x}, e_{\phi}(\mathbf{a_0})]\\
\textbf{h}_1 &= [\mathbf{x}, e_{\phi}(\mathbf{a})],
\label{eq:h}
\end{align}
where $\phi$ is a learnable parameter of the embedding function $e_{\phi}$.

Next, through a GCN, we learn a binary classifier to distinguish between the two datasets. Specifically, for each sample from \( \textbf{h}_0 \) and \( \textbf{h}_1 \), the classifier uses a GCN to aggregate the adjacency information of the neighbors and outputs \( f \in [0, 1]\), representing the probability that the sample comes from the behavior policy. The probability that the sample comes from the evaluation policy is then \( 1 - f \). Let vector \(\textbf{f}_0\) denote the outputs of samples in the historical dataset and let vector \(\textbf{f}_1\) denote the outputs of samples in the dataset where actions are generated by the evaluation policy:
\begin{align}
\textbf{f}_0 &= g_{\theta}(\tilde{A}, \textbf{h}_0) \\
\textbf{f}_1 &= g_{\theta}(\tilde{A}, \textbf{h}_1),
\label{eq:representation_learning}
\end{align}
where $\theta$ is a learnable parameter of GCN $g_{\theta}$ and $\tilde{A} = A + I_n$, since we want the GCN to focus on both the actions of the neighbors and units' own action. To refine the model, we optimize $g_{\theta}$ and $e_{\phi}$ by minimizing the Binary Cross-Entropy (BCE) loss between $\textbf{f}$ and the true label $ L $ of the samples:
\begin{equation}
\mathcal{L} = -\frac{1}{n} \sum_{i=1}^{n} \log(f_{0i}) + \log(1-f_{1i}),
\label{eq:loss}
\end{equation}
where \(f_{0i}\) and \(f_{1i}\) are the \(i\)th values in \(\textbf{f}_0\) and \(\textbf{f}_1\), respectively.

We replace the vanilla importance weights with new weights based on the probabilities given by the classifier:
\begin{equation}
w_i = \frac{f_{0i}}{1 - f_{0i}}.
\label{eq:weight_2}
\end{equation}
Since the weights are assigned to the rewards from the historical dataset, \(\textbf{f}_1\) does not appear in Eq.\ref{eq:weight_2}. It is used for training only. 

These weights are equivalent to the weights of the ideal IPW style estimator mentioned above. The proof is as follows:
\begin{align}
&\quad w(a_i, a_{i\mathcal{N}}, x_i) \notag \\
&= \frac{p(a_i, a_{i\mathcal{N}} | x_i, A, \pi)}{p(a_i, a_{i\mathcal{N}} | x_i, A, \pi_0)} = \frac{p(a_i, a_{i\mathcal{N}} | x_i, A, L=1)}{p(a_i, a_{i\mathcal{N}} | x_i, A, L=0)} \notag  \\
&= \frac{p(a_i, a_{i\mathcal{N}} | x_i, A, L=1) p(x_i, A, L=1) / p(a_i, a_{i\mathcal{N}}, x_i, A)}{p(a_i, a_{i\mathcal{N}} | x_i, A, L=0) p(x_i, A, L=0) / p(a_i, a_{i\mathcal{N}}, x_i, A)} \notag \\
&= \frac{p(L=1 | a_i, a_{i\mathcal{N}}, x_i, A)}{p(L=0 | a_i, a_{i\mathcal{N}}, x_i, A)} = \frac{f_{0i}}{1 - f_{0i}}.
\label{eq:weight_proof}
\end{align}

While maintaining consistency, we reduce the variance by self-normalizing the weights~\cite{NIPS2015_39027dfa}. Finally, we can evaluate a policy in the presence of interference:
\begin{equation}
\hat{V}_{IntIPW} = \frac{\sum_{i=1}^{n} w_i r_i}{\sum_{i=1}^{n} w_i}.
\label{eq:our_method}
\end{equation}
We show that our method is unbiased and that its variance is bounded by
\[
\mathrm{Var}\bigl(\widehat V_{\mathrm{IntIPW}}\bigr)
\le
\frac{1}{n}
\Bigl[(W_{\max}R_{\max})^{2} + W_{\max}^{2}R_{\max}^{2}\Bigr],
\]
where $W_{\max}, R_{\max}$ are the upper bounds of $w$ and $r$, respectively. See Appendix for the proof.
\begin{algorithm}
\caption{IntIPW}
\begin{algorithmic}[1]
    \renewcommand{\algorithmicrequire}{\textbf{Input:}}
    \renewcommand{\algorithmicensure}{\textbf{Output:}}
    \REQUIRE Historical dataset $\mathcal{D} = \{(x_i, a_i, r_i)\}_{i=1}^{n}$ generated by the behavior policy $\pi_0$, new actions \textbf{a} generated by the evaluation policy $\pi$, adjacency matrix $A$, learning rate $\alpha$
    \ENSURE Average estimated reward for all units under new actions
    \STATE $\tilde{A} = A + I_n$
    \FOR{epoch in $\text{max epochs}$}
        \STATE $\textbf{h}_0 \gets [\textbf{x}, e_{\phi}(\textbf{a}_0)]$
        \STATE $\textbf{h}_1 \gets [\textbf{x}, e_{\phi}(\textbf{a})]$
        \STATE $\textbf{f}_0 \gets g_{\theta}(\tilde{A}, \textbf{h}_0)$
        \STATE $\textbf{f}_1 \gets g_{\theta}(\tilde{A}, \textbf{h}_1)$
        \STATE $\mathcal{L} = -\frac{1}{N} \sum_{i=1}^{N} (\log(f_{0i}) + \log(1 - f_{1i}))$
        \STATE Update $\theta \gets \theta - \alpha \frac{\partial \mathcal{L}}{\partial \theta}$
        \STATE Update $\phi \gets \phi - \alpha \frac{\partial \mathcal{L}}{\partial \phi}$
    \ENDFOR
    \STATE \textbf{return} $\frac{\sum_{i=1}^{N} \frac{f_{0i}}{1 - f_{0i}} r_i}{\sum_{i=1}^{N} \frac{f_{0i}}{1 - f_{0i}}}$
\end{algorithmic}
\label{alg:intope}
\end{algorithm}

\begin{table*}[t]
\centering
\caption{The experimental results on the synthetic datasets. The smaller metric means better performance, and the best 
performance of MSE is marked bold.}
\label{tab:synthetic}
\resizebox{1\linewidth}{!}{
\begin{tabular}{@{}lccccccccccccccccccc@{}}
\toprule
\multicolumn{16}{c}{\textit{Erdős-Rényi graph}} \\\hline
& \multicolumn{3}{c}{$ \beta = -1 $} & \multicolumn{3}{c}{$ \beta = -0.75 $} & \multicolumn{3}{c}{$ \beta = -0.5 $} & \multicolumn{3}{c}{$ \beta = -0.25 $} & \multicolumn{3}{c}{$ \beta = 0 $} \\
\cmidrule(r){2-4} \cmidrule(lr){5-7} \cmidrule(lr){8-10} \cmidrule(lr){11-13} \cmidrule(lr){14-16} 
Methods & MSE & Bias & SD & MSE & Bias & SD & MSE & Bias & SD & MSE & Bias & SD & MSE & Bias & SD  \\
\midrule
DM  & 1.425 & 1.130 & 0.386 & 1.410 & 1.122 & 0.390 & 1.353 & 1.100 & 0.377 & 1.180 & 1.023 & 0.365 & 0.194 & 0.420 & 0.133 \\

IPW & 1.429 & 1.131 & 0.386 & 1.413 & 1.123 & 0.389 & 1.356 & 1.102 & 0.376 & 1.184 & 1.026 & 0.364 & 0.193 & 0.419 & 0.133 \\

SNIPW & 1.429 & 1.131 & 0.386 & 1.414 & 1.123 & 0.389 & 1.356 & 1.102 & 0.377 & 1.184 & 1.025 & 0.365 & 0.193 & 0.419 & 0.133 \\

BIPW & 1.532 & 1.187 & 0.352 & 1.516 & 1.179 & 0.355 & 1.457 & 1.157 & 0.344 & 1.273 & 1.074 & 0.344 & 0.229 & 0.455 & 0.148 \\

SGIPW & 1.835 & 1.302 & 0.375 & 1.811 & 1.290 & 0.383 & 1.730 & 1.262 & 0.373 & 1.486 & 1.162 & 0.368 & 0.233 & 0.462 & 0.140 \\

DR    & 1.425 & 1.130 & 0.386 & 1.410 & 1.122 & 0.390 & 1.352 & 1.100 & 0.377 & 1.181 & 1.023 & 0.365 & 0.193 & 0.419 & 0.133 \\

IntIPW & \textbf{1.266} & 1.033 & 0.446 & \textbf{1.287} & 1.049 & 0.431 & \textbf{1.152} & 0.985 & 0.427 & \textbf{1.111} & 0.936 & 0.485 & \textbf{0.163} & 0.383 & 0.128  \\\hline
\multicolumn{16}{c}{\textit{Watts-Strogatz graph}} \\\hline
& \multicolumn{3}{c}{$ \beta = -1 $} & \multicolumn{3}{c}{$ \beta = -0.75 $} & \multicolumn{3}{c}{$ \beta = -0.5 $} & \multicolumn{3}{c}{$ \beta = -0.25 $} & \multicolumn{3}{c}{$ \beta = 0 $} \\
\cmidrule(r){2-4} \cmidrule(lr){5-7} \cmidrule(lr){8-10} \cmidrule(lr){11-13} \cmidrule(lr){14-16} 
Methods & MSE & Bias & SD & MSE & Bias & SD & MSE & Bias & SD & MSE & Bias & SD & MSE & Bias & SD  \\
\midrule
DM  & 1.450 & 1.140 & 0.389 & 1.423 & 1.129 & 0.384 & 1.360 & 1.103 & 0.378 & 1.164 & 1.017 & 0.361 & 0.204 & 0.432 & 0.132 \\

IPW & 1.452 & 1.141 & 0.388 & 1.425 & 1.130 & 0.384 & 1.363 & 1.105 & 0.378 & 1.167 & 1.018 & 0.361 & 0.204 & 0.432 & 0.132 \\

SNIPW & 1.452 & 1.140 & 0.389 & 1.425 & 1.130 & 0.384 & 1.363 & 1.105 & 0.379 & 1.168 & 1.018 & 0.362 & 0.204 & 0.431 & 0.132 \\

BIPW & 1.548 & 1.192 & 0.358 & 1.522 & 1.181 & 0.350 & 1.463 & 1.157 & 0.354 & 1.273 & 1.073 & 0.347 & 0.240 & 0.466 & 0.153 \\

SGIPW & 1.874 & 1.312 & 0.393 & 1.836 & 1.298 & 0.391 & 1.757 & 1.267 & 0.388 & 1.488 & 1.160 & 0.376 & 0.249 & 0.476 & 0.148 \\

DR    & 1.450 & 1.140 & 0.389 & 1.423 & 1.129 & 0.385 & 1.360 & 1.103 & 0.378 & 1.165 & 1.017 & 0.361 & 0.204 & 0.432 & 0.132 \\


IntIPW & \textbf{1.396} & 1.107 & 0.414 & \textbf{1.359} & 1.098 & 0.392 & \textbf{1.330} & 1.089 & 0.378 & \textbf{1.103} & 0.984 & 0.366 & \textbf{0.168} & 0.395 & 0.109 \\
\bottomrule
\end{tabular}
}

\end{table*}

\section{Experiments}
We conducted extensive experiments on both synthetic and real-world datasets to validate the effectiveness of our method IntIPW.

\subsection{Baseline Estimators}

Our baseline estimators are \textbf{Direct Method (DM)}~\cite{beygelzimer2016offset} as described in Eq.\ref{eq:DM}, which learns a predictive reward function from the log data and then uses it to estimate the evaluation policy; \textbf{Inverse Probability Weighting (IPW)}~\cite{Horvitz1952AGO, DBLP:journals/corr/abs-1003-0120}, as described in Eq.\ref{eq:IPW} which re-weights the rewards based on the propensities of the behavior
and evaluation policies;
\textbf{Self Normalized Inverse Probability Weighting (SNIPW)}~\cite{NIPS2015_39027dfa, kallus2019intrinsically} as described in Eq.\ref{eq:SNIPW}, which normalizes the observed rewards by the self-normalized importance weight.
\textbf{Balanced Inverse Probability Weighting (BIPW)}~\cite{sondhi2020balanced}, which re-weights the rewards by the importance weights estimated via a supervised classification procedure; 
\textbf{Sub Gaussian Inverse Probability Weighting (SGIPW)}~\cite{NEURIPS2021_4476b929}, which modifies the vanilla weights by applying power mean; 
\textbf{Doubly Robust (DR)}~\cite{dudik2011doubly} as described in Eq.\ref{eq:dr}, which combines DM and IPW;

We invoked these baseline estimators from the Open Bandit Pipeline~\cite{saito2020open}.

\subsection{Experiment Settings}
\textbf{Units' contexts.} Each unit's context is represented by a 10-dimensional vector. In synthetic experiments, the context vectors are independently Gaussian distributed: 
$x_i \overset{\text{iid}}{\sim} \mathcal{N}(0,1), 1 \leq i \leq 10$.
In real-world datasets, the dimensions of the contexts are very large, so we use Principal Component Analysis (PCA) from scikit-learn~\cite{scikit-learn} to reduce the contexts to 10 dimensions for consistency. \\ 
\textbf{Reward function.} We drew inspiration from~\cite{10.1145/3543507.3583448} to set the reward function as: 
\begin{equation}
\begin{aligned}
r(x_i, a_i, a_{i\mathcal{N}}) &= \big( \beta_1^\top x_i \cdot 1.5^{\beta_2^\top x_i} + b \left| \beta_3^\top x_i + e(a_i)^\top x_i \right|^2  \\ &\quad + c \left| \beta_4^\top x_i + e(a_{i\mathcal{N}})^\top x_i \right|^2 \cdot \sqrt{|N(i)|}  \big) ^ {\frac{1}{3}} + \varepsilon_i,
\end{aligned}
\label{eq:reward_function}
\end{equation}
where $\beta_1$, $\beta_2$, $\beta_3$, $\beta_4$ are 10-dimensional vectors independently generated from Gaussian distribution: $\beta_i \overset{\text{iid}}{\sim} \mathcal{N}(0,1), 1 \leq i \leq 4$; $e(a)$ is an embedding function that maps actions into the semantic space; $|N(i)|$ is the number of neighbors of unit $i$; $\varepsilon_i \overset{\text{iid}}{\sim} \mathcal{N}(0,1)$ is the error term of unit $i$; $b$ and $c$, both with default values of 1, are coefficients that control the impact of one's own action and the impact of the neighbors' actions, respectively. $a_\mathcal{N}$ is set to be the linear aggregation of the neighbors' actions. For instance, when the action is binary, $a_\mathcal{N}$ can be interpreted as a value in $ [0,1] $, representing the proportion of unit $i$'s neighbors having action = 1. Correspondingly, $e(a_{i\mathcal{N}})$ denotes the influence of the linear aggregation of neighbors' actions $a_{i\mathcal{N}}$ on unit $i$:
\begin{equation}
e(a_{i\mathcal{N}}) = \frac{\sum_{j \in N(i)} e(a_j)}{|N(i)|}.
\label{eq:e(ain)}
\end{equation}
\textbf{Behavior policy.} Following~\cite{saito2022offpolicy}, our behavior is a function similar to the softmax function:
\begin{equation}
\pi_0(a_i | x_i) = \frac{exp(r_{\text{direct}}(x_i,a_i)\beta)}{\sum_{a_j \in A} exp(r_{\text{direct}}(x_i,a_j)\beta)},
\label{eq:behavior_policy}
\end{equation}
where $ r_{\text{direct}} $ is a function similar to Eq.\ref{eq:reward_function} but does not take into account the influence of the neighbors' actions:
\begin{equation}
r_{\text{direct}}(x_i, a) = \beta_1^\top x_i \cdot 1.5^{\beta_2^\top x_i} + \left| \beta_3^\top x_i + e(a_{i\mathcal{N}})^\top x_i \right|^2,
\label{eq:r_direct}
\end{equation}
and $\beta$ acts as an exploration-exploitation controller. 
A larger $\beta$ implies that the action with the greatest $r_{\text{direct}}$ is more likely to be selected.
It is worth noting that when $\beta$ equals zero, $ \pi_0 $ is a purely random strategy, with each action having an equal probability of being selected. \\
\textbf{Evaluation policy.} The evaluation policy is similar to that in~\cite{10.1145/3543507.3583448}:
\begin{equation}
\pi(a_i|x_i) = \begin{cases} 
      \gamma & \text{if } a_i = \arg\max_{a' \in A},[r_{\text{direct}}(x_i,a')],\ \\
      \frac{1 - \gamma}{D-1} & \text{otherwise,}
   \end{cases}
\label{eq:evaluate_a_policy}
\end{equation}
where \(D\) is the number of actions and $\gamma \in [0, 1]$ is the probability that the action with the greatest $ r_{\text{direct}} $ is selected. Other actions are selected with an equal probability of $\frac{1-\gamma}{D-1}$. We set $\gamma$ to 0.8. \\
\textbf{Validation metrics.}
In off-policy evaluation, bias measures how well an estimator handles shifts in distributions. However, due to the impracticality of frequently sampling actions from the behavior policy for multiple off-policy evaluations, the MSE is a more crucial and representative metric~\cite{10.1145/3543507.3583448}. We have also argued previously in Eq.\ref{eq:mse_decomposed} that MSE can be considered a combination of the capability to handle distribution shifts (bias) and stability (standard deviation).

\begin{table*}[t]
\centering
\caption{The experimental results on the real-world datasets. The smaller metric means better performance, and the best 
performance of MSE is marked bold.}
\label{tab:real_world}
\resizebox{1\linewidth}{!}{
\begin{tabular}{@{}lccccccccccccccccccc@{}}
\toprule
\multicolumn{16}{c}{\textit{BlogCatalog}} \\\hline
& \multicolumn{3}{c}{$ \beta = -1 $} & \multicolumn{3}{c}{$ \beta = -0.75 $} & \multicolumn{3}{c}{$ \beta = -0.5 $} & \multicolumn{3}{c}{$ \beta = -0.25 $} & \multicolumn{3}{c}{$ \beta = 0 $} \\
\cmidrule(r){2-4} \cmidrule(lr){5-7} \cmidrule(lr){8-10} \cmidrule(lr){11-13} \cmidrule(lr){14-16} 
Methods & MSE & Bias & SD & MSE & Bias & SD & MSE & Bias & SD & MSE & Bias & SD & MSE & Bias & SD  \\
\midrule
DM & 10.729 & 2.860 & 1.596 & 9.764 & 2.731 & 1.518 & 9.101 & 2.704 & 1.338 & 7.286 & 2.417 & 1.201 & 4.561 & 1.888 & 0.998  \\

IPW & 12.815 & 2.052 & 2.933 & 11.488 & 2.630 & 2.138 & 9.353 & 2.423 & 1.866 & 6.778 & 2.276 & 1.263 & 4.205 & 1.778 & 1.022 \\

SNIPW & 11.659 & 2.166 & 2.640 & 10.829 & 2.580 & 2.043 & 8.894 & 2.387 & 1.788 & 6.629 & 2.247 & 1.258 & 4.203 & 1.778 & 1.021  \\

BIPW & 14.621 & 2.811 & 2.592 &  13.287 & 2.677 & 2.474 & 11.929 & 2.705 & 2.148 & 9.988 & 2.547 & 3.872 & 6.316 & 2.152 & 1.299  \\

SGIPW & 29.856 & 4.917 & 2.383 & 26.301 & 4.606 & 2.254 & 22.703 & 4.326 & 1.997 & 16.542 & 3.703 & 1.682 & 8.479 & 2.635 & 1.239 \\

DR & 15.806 & 3.227 & 2.322 &  10.946 & 2.831 & 1.712 & 8.799 & 2.586 & 1.453 & 6.698 & 2.262 & 1.258 & 4.275 & 1.780 & 1.053  \\

IntIPW & \textbf{9.595} & 2.555 & 1.751 & \textbf{7.438} & 2.298 & 1.469 & \textbf{5.575} & 2.091 & 1.097 & \textbf{4.299} & 1.776 & 1.070 & \textbf{2.317} & 1.192 & 0.947  \\\hline
\multicolumn{16}{c}{\textit{Flickr}} \\\hline
& \multicolumn{3}{c}{$ \beta = -1 $} & \multicolumn{3}{c}{$ \beta = -0.75 $} & \multicolumn{3}{c}{$ \beta = -0.5 $} & \multicolumn{3}{c}{$ \beta = -0.25 $} & \multicolumn{3}{c}{$ \beta = 0 $} \\
\cmidrule(r){2-4} \cmidrule(lr){5-7} \cmidrule(lr){8-10} \cmidrule(lr){11-13} \cmidrule(lr){14-16} 
Methods & MSE & Bias & SD & MSE & Bias & SD & MSE & Bias & SD & MSE & Bias & SD & MSE & Bias & SD  \\
\midrule
DM & 3.237 & 0.906 & 1.554 & 2.896 & 0.871 & 1.462 & 2.139 & 0.562 & 1.350 & 1.827 & 0.509 & 1.25 & 1.370 & 0.395 & 1.102  \\
IPW & 4.115 & 0.623 & 1.931 & 3.215 & 0.790 & 1.610 & 2.907 & 0.378 & 1.663 & 2.060 & 0.498 & 1.346 & 1.599 & 0.402 & 1.199  \\
SNIPW & 4.021 & 0.673 & 1.889 & 3.226 & 0.793 & 1.612 & 2.875 & 0.387 & 1.651 & 2.049 & 0.492 & 1.344 & 1.596 & 0.401 & 1.198  \\
BIPW & 2.915 & 0.691 & 1.561 & 2.551 & 0.677 & 1.447 & 1.457 & 0.424 & 1.130 & 1.337 & 0.461 & 1.061 & 1.171 & 0.500 & 0.960  \\
SGIPW & 5.777 & 1.727 & 1.671 & 5.227 & 1.645 & 1.588 & 3.530 & 1.268 & 1.386 & 3.049 & 1.149 & 1.315 &  2.275 & 0.912 & 1.202  \\
DR & 3.865 & 0.695 & 1.839 & 5.729 & 0.306 & 2.374 & 1.568 & 0.300 & 1.216 & 1.452 & 0.365 & 1.148 & 1.489 & 0.370 & 1.163  \\

IntIPW & \textbf{1.790} & 0.996 & 0.893 & \textbf{1.318} & 0.862 & 0.758 & \textbf{1.099} & 0.748 & 0.734 & \textbf{1.154} & 0.857 & 0.648 & \textbf{0.851} & 0.691 & 0.611 \\
\bottomrule
\end{tabular}
}
\end{table*}

\subsection{Synthetic Experiments}
\subsubsection{\textit{Synthetic Datasets.}} We choose two random graphs, generated using the Python package \texttt{NetworkX}~\cite{hagberg2008exploring}, to serve as the adjacency matrices. \\
\textbf{Erdős-Rényi (ER) graph}~\cite{ErdosRenyi1959, Gilbert1959}. In this model, a graph is constructed by connecting nodes completely at random. Each pair of units has a fixed probability $p$ of becoming neighbors.\\
\textbf{Watts-Strogatz graph}~\cite{Watts1998}. The construction of a Watts–Strogatz graph starts with a regular ring lattice, where each node is connected to its $k$ neighbors. In our experiment, we arranged the units according to the results of reducing the contexts to one dimension using PCA. Then, each edge is randomly rewired, allowing for long-range connections.\\
\indent Although these two random graphs not necessarily reflect real-world situations, their advantage is that we can freely change the sample size and the number of neighbors to study their impact on the results.

\subsubsection{\textit{Results.}} The default settings for the experiment are as follows: sample size $n = 10,000$; each unit has 10 neighbors on average, that is, $p = \frac{10}{n}$ for Erdős-Rényi graph and $k = 10$ for Watts-Strogatz graph; actions are binary; $\beta$ in the behavior policy Eq.\ref{eq:behavior_policy} is set to 0, meaning the behavior policy is purely random; $\gamma$ in the evaluation policy Eq.\ref{eq:evaluation_policy} is set to 0.8, indicating that the evaluation policy has an 0.8 probability of choosing the optimal action for the unit itself; $b$ and $c$ in the reward function Eq.\ref{eq:reward_function} are both set to 1, signifying that the weights of the impact of one's own action and the neighbors' actions are the same. We modify these default settings to compare the performance of our IntIPW estimator with baseline estimators. For each setting, we conduct experiments with 20 different seeds. 

\textbf{How does IntIPW perform under different selection bias strength $\beta$ of the behavior policy?}
We set $\beta$ in [-1, -0.75, -0.5, -0.25, 0]. A higher $\beta$ value means that each unit in the behavior policy has a higher probability of choosing the optimal action for itself (the action with the greatest value of $ r_{\text{direct}} $), which also implies less discrepancy between the behavior and evaluation policies. As demonstrated in Tab.\ref{tab:synthetic}, the results of the two graphs are very close. When the discrepancy between the behavior and evaluation policies decreases, all estimators show improved performance, with reductions across all three metrics. For both graphs, our method not only has the smallest MSE but also the least bias across all experiments. 

\begin{figure}[h]
\centering
\includegraphics[width=1\linewidth]{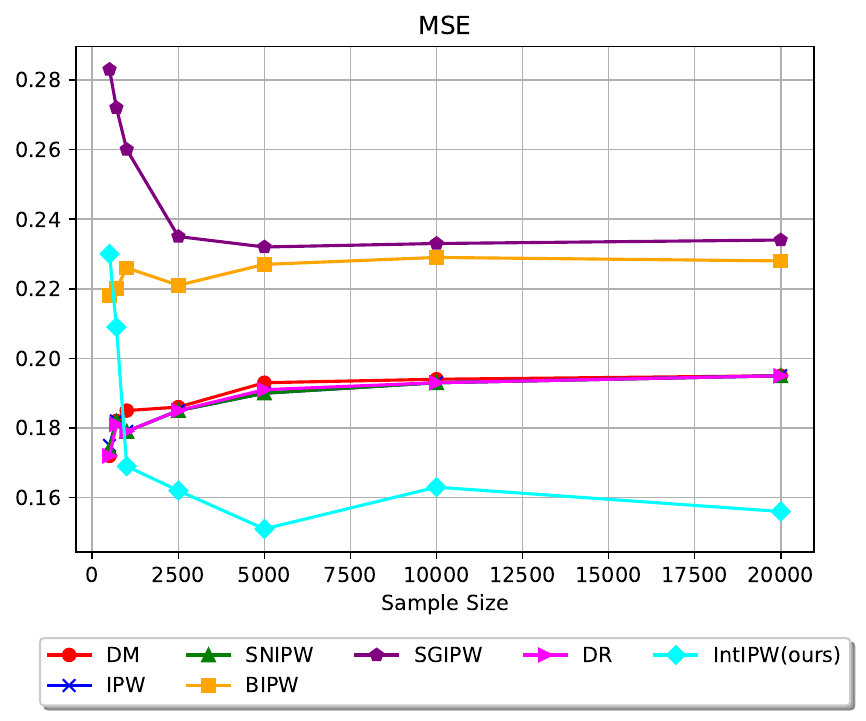}
\caption{Performance of estimators in different sample sizes}
\label{fig:fig2_sample_size}
\end{figure}

\textbf{How does IntIPW perform in different sample sizes?}
We keep the setting that each unit has 10 neighbors unchanged and set the sample size $N$ in [500, 700, 1,000, 2,500, 5,000, 10,000, 20,000]. We conduct the experiment with the Erdős-Rényi graph. As results shown in Fig.\ref{fig:fig2_sample_size}, when the sample size is too small (e.g., $N = 500$), the performance of IntIPW is not satisfactory. This is likely because, compared to other estimators, IntIPW employs a GCN to aggregate the influence of neighbors, which introduces complexity into the model. Consequently, a larger sample size is required for effective training of this model. 
As the sample size increases, the MSE of all estimators tends to stabilize, and our method becomes the optimal one. \\

\begin{figure}[ht]
\centering
\includegraphics[width=1\linewidth]{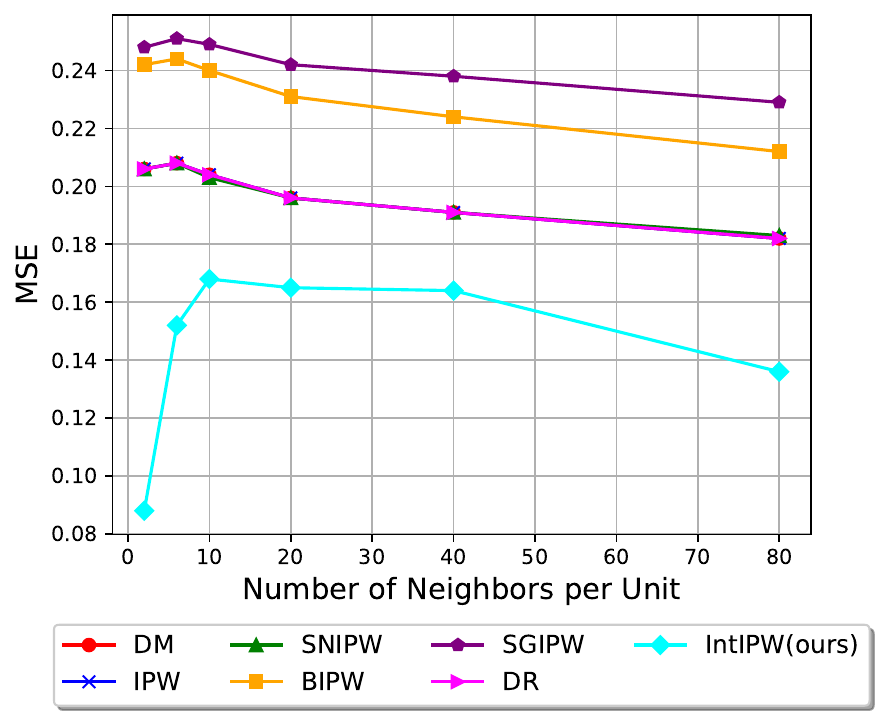}
\caption{Performance of estimators with varying average numbers of neighbors}
\label{fig:fig3_neighbor}

\end{figure}
\textbf{How does IntIPW perform with different number of neighbors?}
We keep the sample size $n$ = 10,000 unchanged and make each unit have [2, 6, 10, 20, 40, 80] neighbors. We conduct the experiment with the Watts-Strogatz graph. As results demonstrated in Fig.\ref{fig:fig3_neighbor}, 
IntIPW performs very well when the number of neighbors is either very large or very small. The MSE of IntIPW is at its worst when each unit has 10 neighbors (which is the default setting for other experiments), yet even in this scenario, it still outperforms baseline estimators.


\begin{figure*}[!htbp]
\centering
\includegraphics[width=1\linewidth]{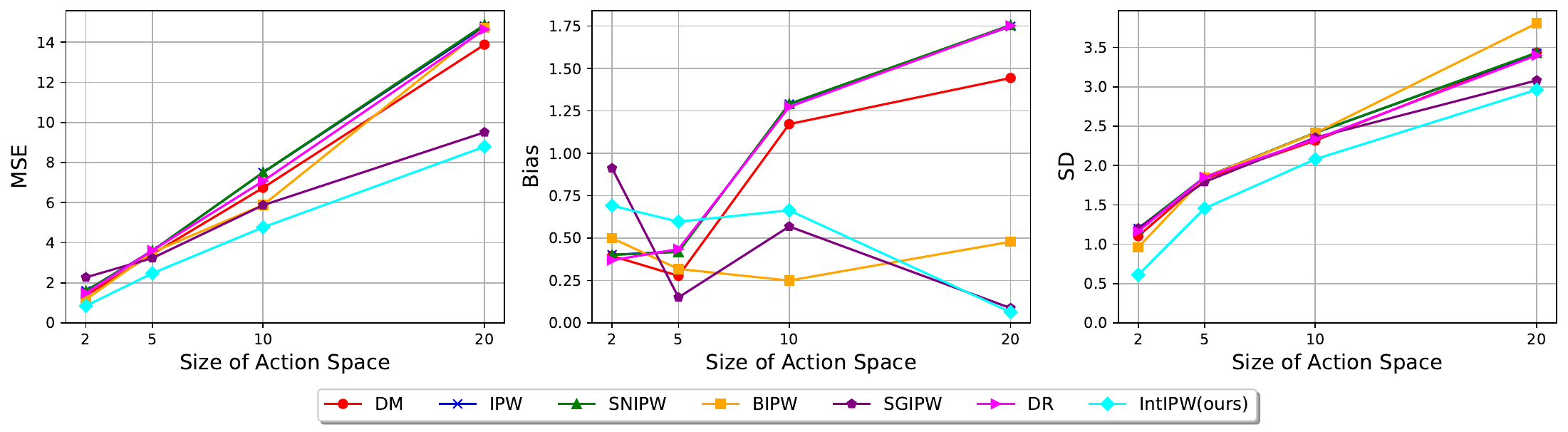}
\caption{Performance of estimators in different action spaces}
\label{fig:fig5_action_space}
\end{figure*}

\subsection{Real-World Experiments}
\subsubsection{\textit{Real-World Datasets.}}  
We use the following two real-world datasets for our experiments.
\newline
\textbf{BlogCatalog}~\cite{10.1145/2898361}. BlogCatalog is a social blog directory that has been used as a dataset for network analysis and social media research. It includes data about bloggers and their connections, organized in a way that represents the social network among them. The dataset we used contains 5,196 units and 171,743 edges. Each unit's context is a 2,246-dimensional vector.\newline
\textbf{Flickr}~\cite{10.1145/2898361}.  Flickr is a web-based platform for storing, organizing, and sharing digital photographs. This network dataset is a collection of data extracted from the Flickr social network, primarily focusing on the connections and interactions among Flickr users. The dataset consists of units representing individual users and edges indicating the follower networks between them. It contains 7,575 units and 239,738 edges. Each unit's context is a 1,205-dimensional vector. 

These two datasets contain the context and adjacency information for each unit, but do not include actions and rewards. Like the synthetic experiments, we use the behavior policy Eq.\ref{eq:behavior_policy}, evaluation policy Eq.\ref{eq:evaluation_policy}, and reward function Eq.\ref{eq:reward_function} specified in the experiment settings.

\subsubsection{\textit{Results.}} 
Similar to the synthetic experiments, the default settings for the experiment are as follows: actions are binary; $\beta$ in the behavior policy Eq.\ref{eq:behavior_policy} is set to 0; $\gamma$ in the evaluation policy Eq.\ref{eq:evaluation_policy} is set to 0.8; $b$ and $c$ in the reward function Eq.\ref{eq:reward_function} are both set to 1. We modify these default settings to compare the performance of our IntIPW estimator with baseline estimators. For each setting, we conduct experiments with 20 different seeds. 

\textbf{How does IntIPW perform under different selection bias strength $\beta$ of the behavior policy?} 
Identical to the synthetic experiment, we set $\beta$ in $[-1, -0.75, -0.5, -0.25, 0]$. 
As can be seen from Tab.\ref{tab:real_world}, as the discrepancy between the behavior and evaluation policies decreases, all estimators perform better, with IntIPW consistently being the best among them. In BlogCatalog, the bias and SD of IntIPW are almost always the lowest as well (the only exception is when $\beta$ = -1). In Flickr, compared with the baseline estimators, IntIPW balances a slight additional bias with a significant reduction in SD, thereby consistently achieving better MSE performance. 

\begin{figure*}[!htbp]
\centering
\includegraphics[width=1\linewidth]{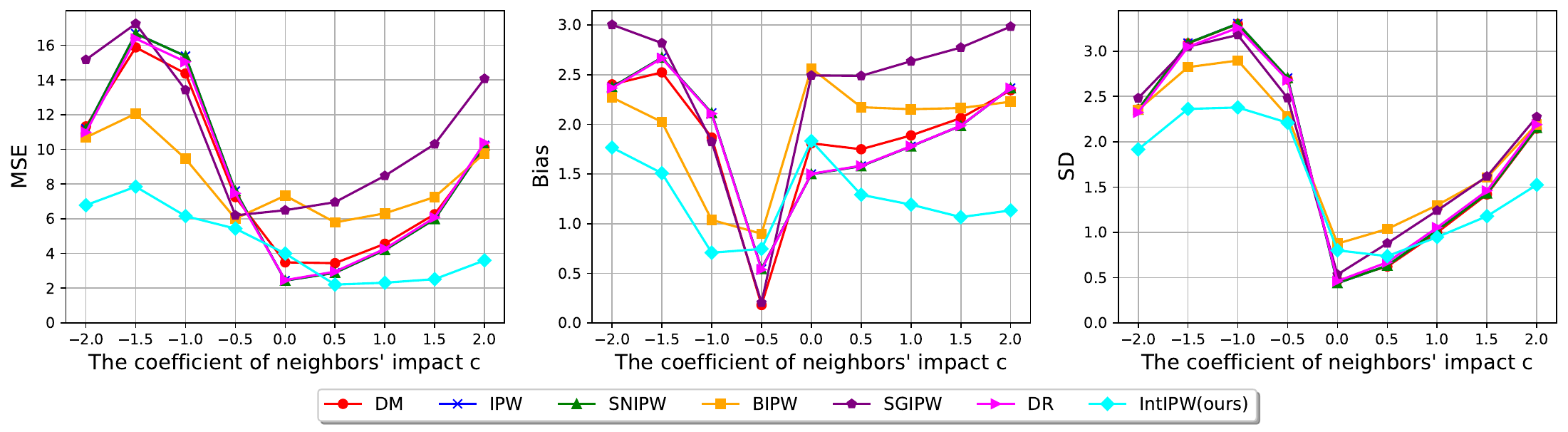}
\caption{Performance of estimators under different intensities of impact of neighbors' actions }
\label{fig:fig4_neighbor_impact}
\end{figure*}

\textbf{When the weight of the impact of one's own action and the impact of the neighbors' actions change, how effective is IntIPW estimator?}
We conducted experiments on BlogCatalog by changing the values of $ b $ and $ c $ in the reward function Eq.\ref{eq:reward_function}. We selected 9 $ (b, c) $ value pairs: $ (0, -2) $, $ (0.5, -1.5) $, $ (1, -1) $, $ (1.5, -0.5) $, $ (2, 0) $, $ (1.5, 0.5) $, $ (1, 1) $, $ (0.5, 1.5) $, and $ (0, 2) $. The results are shown in Fig.\ref{fig:fig4_neighbor_impact}. From the MSE  graph, we can see that when the neighbors have no influence (i.e., $c = 0$), our method is slightly inferior to the baseline estimators. This is expected, as baseline estimators only consider the impact of one's own action, making them well-suited for this setting. As the absolute value of $c$ increases, the MSE of baseline estimators rapidly increases, making our method the best. This is because they only focus on the impact of actions taken by the unit itself, completely ignoring the influence of neighbors.

\textbf{When the action space is large, does IntIPW estimator still perform well?} 
We vary the number of actions $D$ in [2, 5, 10, 20] on Flickr. As results demonstrated in Fig.\ref{fig:fig5_action_space}, as the action space gets augmented, the performance of all estimators worsens, but IntIPW consistently has the smallest MSE and SD. When $D = 20$, our method also has the smallest bias. This indicates that our method is robust across different action spaces.

\section{Conclusion and Future Work}

In this paper, our goal is to address the problem of OPE in the presence of interference. We point out that existing estimators are biased in this setting because they do not consider interference. To address this issue, we propose a new estimator, IntIPW. It trains a classifier through a GCN to give the probabilities of a unit coming from the behavior and evaluation policies, and then assigns weights to the samples in the historical dataset based on the ratio of these probabilities. Extensive results on both synthetic and real-world datasets prove the effectiveness and robustness of IntIPW.

Our work raises several questions for further research. First, our current method is only applicable to discrete action spaces, so how to generalize our method to continuous action spaces is a direction worth exploring. In addition, in our experiment setting, the influence of actions of neighbors is homogeneous, which is often not true in real life~\cite{qu2022efficient, song2019session, ma2011recommender}. Whether our method remains effective in real-world scenarios with heterogeneous interference is a matter worthy of investigation.

\bibliographystyle{IEEEtran}
\bibliography{intope} 

\newpage
\appendix
Recall that we let $\mathcal{G}=(V,E)$ be an undirected graph on $n$ units, $V=\{1,\dots,n\}$. For each unit $i$ we observe:
\begin{itemize}
  \item a context (feature) vector $x_{i}\in\mathbb{R}^{p}$,
  \item an action $a_{i}\in\{1,\dots,D\}$ chosen under the logging policy $\pi_{0}$,
  \item a reward $r_{i}\in\mathbb{R}$.
\end{itemize}
Denote by $N(i)\subset V$ the neighborhood of $i$ in $\mathcal{G}$, and let
\[
a_{\,iN}\;:=\;\{\,a_{j}: j\in N(i)\}.
\]
We write $A=(a_{1},\dots,a_{n})$ for the vector of all actions.  Define the ``potential outcome’’
$Y_{i}(a_{i},\,a_{\,iN})\;=\;$
the reward of unit $i$ if it and its neighbors take $(a_{i},\,a_{\,iN})$.

We aim to evaluate a deterministic target policy $\pi$ which for each $i$ and context $x_{i}$ prescribes 
\[
\pi_{i}(x_{i}) \;=\; \bigl(\tilde a_{i},\,\tilde a_{\,iN}\bigr).
\]
Under this policy, the \emph{policy value} is
\[
V(\pi)
\;=\;
\frac{1}{n}\sum_{i=1}^{n}
\mathbb{E}\bigl[Y_{i}\bigl(\pi_{i}(x_{i})\bigr)\bigr],
\]
where the expectation is over any randomness in $x_{i}$ and in how a unit’s outcome is generated given $(a_{i},a_{\,iN})$.

In the logged data, each $(x_{i},a_{i},a_{\,iN},r_{i})$ is drawn under $\pi_{0}$, so that
\[
r_{i}
\;\bigl|\;(x_{1:n},\,A)\;\overset{\mathrm{d}}{=}
\;r_{i}\;\bigl|\;(x_{i},\,a_{i},\,a_{\,iN}\bigr).
\]
Define the \emph{oracle importance weight}
\[
\begin{aligned}
  w_{i}^{*} \;&:=\; 
    \frac{\pi\bigl(a_{i},\,a_{\,iN}\mid x_{i},\,A\bigr)}
         {\pi_{0}\bigl(a_{i},\,a_{\,iN}\mid x_{i},\,A\bigr)} 
  \\
  \;&=\;
  \begin{cases}
    \displaystyle \frac{1}{\pi_{0}\bigl(a_{i},\,a_{\,iN}\mid x_{i},\,A\bigr)}
      & \text{if}\;(a_{i},\,a_{\,iN}) = \pi_{i}(x_{i}), \\
    0
      & \text{otherwise.}
  \end{cases}
\end{aligned}
\]

The \emph{self‐normalized IntIPW estimator} is
\[
\widehat V_{\mathrm{IntIPW}}
\;=\;
\frac{\sum_{i=1}^{n} w_{i}^{*}\,r_{i}}{\sum_{i=1}^{n} w_{i}^{*}}.
\]
To analyze bias, variance, and consistency when replacing $w_{i}^{*}$ by an estimate $\widehat w_{i}$, we state the following assumptions:

\begin{assumption}[Local‐Interference / Markov Independence]
For each $i$, the distribution of $r_{i}$ given $(x_{1:n},\,A)$ depends only on $(x_{i},\,a_{i},\,a_{\,iN})$, i.e.
\[
r_{i}\;\bigl|\;(x_{1:n},\,A)\;\overset{\mathrm{d}}{=}\;\bigl.r_{i}\;\bigl|\;(x_{i},\,a_{i},\,a_{\,iN}\bigr).
\]
\end{assumption}

\begin{assumption}[Positivity / Overlap]
There exists $\alpha>0$ such that for all $i$ and all $(x_{i},a_{i},a_{\,iN})$ in the support,
\[
\pi_{0}(a_{i},\,a_{\,iN}\mid x_{i},\,A)\;\ge\;\alpha.
\]
\end{assumption}

\begin{assumption}[Bounded Outcomes]
There is a constant $R_{\max}<\infty$ such that for all $i$ and any $(x_{i},a_{i},a_{\,iN})$,
\[
|r_{i}|\;\le\;R_{\max}.
\]
\end{assumption}

\begin{assumption}[Oracle Weights Bounded]
There is $W_{\max}<\infty$ such that
\[
0 \;\le\; w_{i}^{*}\;\le\; W_{\max}
\quad\text{a.s.,}
\]
hence in particular $w_{i}^{*}\le 1/\alpha$ by Assumption 2.
\end{assumption}

\section{Theorem 1: Bias and Variance with Oracle Weights}

\begin{theorem}
Under Assumptions 1–4, set $\widehat w_{i}=w_{i}^{*}$ exactly.  Then the self‐normalized IntIPW estimator
\[
\widehat V_{\mathrm{IntIPW}}
\;=\;
\frac{\sum_{i=1}^{n} w_{i}^{*}\,r_{i}}{\sum_{i=1}^{n} w_{i}^{*}}
\]
satisfies:
\begin{enumerate}
  \item \textbf{Unbiasedness:}
    \[
      \mathbb{E}\bigl[\widehat V_{\mathrm{IntIPW}}\bigr]
      \;=\;
      V(\pi).
    \]
  \item \textbf{Variance:}
    \[
      \mathrm{Var}\bigl(\widehat V_{\mathrm{IntIPW}}\bigr)
      \;=\;
      \frac{1}{n\bigl(\mathbb{E}[\,w_{1}^{*}\bigr])^{2}}\,
      \mathrm{Var}\bigl(w_{1}^{*}\,r_{1}\bigr)
      \;-\;
      \frac{1}{n}
      \,\frac{\mathrm{Var}(w_{1}^{*})}{\bigl(\mathbb{E}[\,w_{1}^{*}\bigr])^{2}}\,
      \bigl(\mathbb{E}[\,r_{1}\,]\bigr)^{2}.
    \]
    In particular, since $|r_{1}|\le R_{\max}$ and $0\le w_{1}^{*}\le W_{\max}$,
    \[
      \mathrm{Var}\bigl(\widehat V_{\mathrm{IntIPW}}\bigr)
      \;\le\;
      \frac{1}{n}
      \Bigl[
        (W_{\max}R_{\max})^{2} \;+\; W_{\max}^{2}R_{\max}^{2}
      \Bigr].
    \]
\end{enumerate}
\end{theorem}

\begin{proof}
\textbf{(1) Unbiasedness.}
For each $i$, let
\[
\mu_{i}(x_{i},a_{i},a_{\,iN})
\;=\;
\mathbb{E}[\,r_{i}\mid x_{i},\,a_{i},\,a_{\,iN}\bigr].
\]
Under Assumptions 1–2, the logged data $(x_{i},a_{i},a_{\,iN},r_{i})$ are i.i.d.\ under $\pi_{0}$.  Then
\begin{align*}
\mathbb{E}\bigl[w_{i}^{*}\,r_{i}\bigr]
&=\;\mathbb{E}_{x_{i},A}\;
   \mathbb{E}_{(a_{i},a_{\,iN},r_{i})\mid(x_{i},A)}
   \bigl[w_{i}^{*}\,r_{i}\bigr]\\
&=\;\mathbb{E}_{x_{i},A}\;
   \sum_{(a_{i},a_{\,iN})}\;
   \pi_{0}\bigl(a_{i},\,a_{\,iN}\mid x_{i},A\bigr)\,
   \bigl[w_{i}^{*}\,\mu_{i}(x_{i},a_{i},a_{\,iN})\bigr].
\end{align*}
Since $w_{i}^{*}= \pi(a_{i},a_{\,iN}\mid x_{i},A)\,/\,\pi_{0}(a_{i},a_{\,iN}\mid x_{i},A)$ whenever $(a_{i},a_{\,iN})=\pi_{i}(x_{i})$, and $0$ otherwise, it follows that
\[
\pi_{0}(a_{i},\,a_{\,iN}\mid x_{i},A)\;w_{i}^{*}
=\;
\pi\bigl(a_{i},\,a_{\,iN}\mid x_{i},A\bigr)\,
\mathbf{1}\{(a_{i},a_{\,iN})=\pi_{i}(x_{i})\}.
\]
Hence
\[
\mathbb{E}\bigl[w_{i}^{*}\,r_{i}\bigr]
=\;\mathbb{E}_{x_{i},A}\;
   \mu_{i}\bigl(x_{i},\pi_{i}(x_{i})\bigr)
=\;\mathbb{E}\bigl[Y_{i}(\pi_{i}(x_{i}))\bigr].
\]
Summing over $i=1,\dots,n$ gives
\[
\sum_{i=1}^{n}\mathbb{E}\bigl[w_{i}^{*}\,r_{i}\bigr]
=\;n\,V(\pi).
\]
On the other hand,
\[
\mathbb{E}[\,w_{i}^{*}\,]
=\;\mathbb{E}_{x_{i},A}
   \sum_{(a_{i},a_{\,iN})}
   \pi_{0}(a_{i},\,a_{\,iN}\mid x_{i},A)\,w_{i}^{*}
=\mathbb{E}\bigl[1\bigr]=1.
\]
Therefore
\[
\mathbb{E}\bigl[\widehat V_{\mathrm{IntIPW}}\bigr]
=\;\mathbb{E}\Bigl[
   \frac{\sum_{i}w_{i}^{*}\,r_{i}}{\sum_{i}w_{i}^{*}}
   \Bigr]
=\;
\frac{\sum_{i}\mathbb{E}[\,w_{i}^{*}r_{i}\bigr]}{\sum_{i}\mathbb{E}[\,w_{i}^{*}\bigr]}
=\;
\frac{n\,V(\pi)}{n\cdot 1}
=V(\pi).
\]

\medskip

\noindent\textbf{(2) Variance.}
Define
\[
S_{n}=\sum_{i=1}^{n}w_{i}^{*}\,r_{i}, 
\quad
T_{n}=\sum_{i=1}^{n}w_{i}^{*}.
\]
Then $\widehat V_{\mathrm{IntIPW}}=S_{n}/T_{n}$.  Note that $\{(w_{i}^{*},r_{i})\}$ are i.i.d.\ pairs.  Let
\[
\mu_{wr}=\mathbb{E}\bigl[w_{1}^{*}\,r_{1}\bigr]=\mathbb{E}[Y_{1}(\pi_{1}(x_{1}))]=V(\pi),
\quad
\mu_{w}=\mathbb{E}\bigl[w_{1}^{*}\bigr]=1,
\]
\[
\sigma_{wr}^{2}=\mathrm{Var}(w_{1}^{*}\,r_{1}),
\quad
\sigma_{w}^{2}=\mathrm{Var}(w_{1}^{*}),
\quad
\mu_{r}=\mathbb{E}[r_{1}].
\]
By a bivariate CLT and the delta‐method for the ratio of means, for large $n$,
\[
\sqrt{n}
\begin{pmatrix}
\;S_{n}/n - \mu_{wr}\;\\[0.5em]
\;T_{n}/n - \mu_{w}\;
\end{pmatrix}
\;\xrightarrow{d}\;
\mathcal{N}\Bigl(0,\,
\begin{pmatrix}
\sigma_{wr}^{2} & \mathrm{Cov}(w_{1}^{*}r_{1},\,w_{1}^{*})\\
\mathrm{Cov}(w_{1}^{*}r_{1},\,w_{1}^{*}) & \sigma_{w}^{2}
\end{pmatrix}
\Bigr).
\]
Since $\mu_{w}=1$ and $\mu_{wr}=V(\pi)$, the delta‐method yields
\[
\sqrt{n}\bigl(\widehat V_{\mathrm{IntIPW}} - V(\pi)\bigr)
\;\xrightarrow{d}\;
\mathcal{N}\bigl(0,\tau^{2}\bigr),
\]
with
\[
\tau^{2}
=\sigma_{wr}^{2}
-2\,V(\pi)\,\mathrm{Cov}(w_{1}^{*}r_{1},\,w_{1}^{*})
+V(\pi)^{2}\,\sigma_{w}^{2}.
\]
Equivalently, the finite‐sample variance (dropping $O(n^{-2})$ terms) is
\[
\mathrm{Var}\bigl(\widehat V_{\mathrm{IntIPW}}\bigr)
=\frac{1}{n}\,\mathrm{Var}(w_{1}^{*}r_{1})
-\frac{1}{n}\,\mathrm{Var}(w_{1}^{*})\,(\mathbb{E}[r_{1}])^{2}.
\]
Since $|w_{1}^{*}r_{1}|\le W_{\max}R_{\max}$ and $|w_{1}^{*}|\le W_{\max}$,
\[
\mathrm{Var}(w_{1}^{*}r_{1})\le (W_{\max}R_{\max})^{2},
\qquad
\mathrm{Var}(w_{1}^{*})\le W_{\max}^{2}.
\]
Thus
\[
\mathrm{Var}\bigl(\widehat V_{\mathrm{IntIPW}}\bigr)
\le
\frac{1}{n}
\Bigl[(W_{\max}R_{\max})^{2} + W_{\max}^{2}R_{\max}^{2}\Bigr],
\]
completing the proof.
\end{proof}

\section{Theorem 2: Consistency and Asymptotic Normality with Estimated Weights}

\end{document}